\documentclass[sigconf, nonacm]{acmart}

\AtBeginDocument{%
  \providecommand\BibTeX{{%
    Bib\TeX}}}

\usepackage{subcaption}
\usepackage{tablefootnote}
\usepackage{multirow}
\usepackage[normalem]{ulem}
\useunder{\uline}{\ul}{}

 \newcommand{\quotes}[1]{``#1''}

\begin{document}

\title{Stress Detection from Photoplethysmography in a Virtual Reality Environment}

\author{Athar Mahmoudi-Nejad}
\email{athar1@ualberta.ca}
\affiliation{%
  \institution{University of Alberta}
  \city{Edmonton}
  \state{Alberta}
  \country{Canada}
}

\author{Pierre Boulanger}
\email{pierreb@ualberta.ca}
\affiliation{%
  \institution{University of Alberta}
  \city{Edmonton}
  \state{Alberta}
  \country{Canada} 
 }

\author{Matthew Guzdial}
\email{guzdial@ualberta.ca}
\affiliation{%
  \institution{University of Alberta}
  \city{Edmonton}
  \state{Alberta}
  \country{Canada}
}

\renewcommand{\shortauthors}{Mahmoudi-Nejad et al.}

\begin{abstract}
 Personalized virtual reality exposure therapy is a therapeutic practice that can adapt to an individual patient, leading to better health outcomes. Measuring a patient's mental state to adjust the therapy is a critical but difficult task. Most published studies use subjective methods to estimate a patient's mental state, which can be inaccurate. This article proposes a virtual reality exposure therapy (VRET) platform capable of assessing a patient's mental state using non-intrusive and widely available physiological signals such as photoplethysmography (PPG). In a case study, we evaluate how PPG signals can be used to detect two binary classifications: peaceful and stressful states. Sixteen healthy subjects were exposed to the two VR environments (relaxed and stressful). Using LOSO cross-validation, our best classification model could predict the two states with a 70.6\% accuracy which outperforms many more complex approaches.
\end{abstract}

\keywords{Virtual Reality, Stress/Anxiety detection, Physiological measures, Photoplethysmography}

\ccsdesc[500]{Human-centered computing~User studies}
\ccsdesc[500]{Computing methodologies~Classification and regression trees}
\maketitle

\section{Introduction and Related Work}

Virtual Reality (VR) applications are effective tools in treating anxiety-related problems~\cite{maples2017use}. Virtual reality exposure therapy (VRET) is included among such applications~\cite{carl2019virtual}, wherein an individual is immersed in a computer-generated virtual environment to directly confront feared situations or locations that may not be practical or safe to encounter in real life.   

Existing VR applications have attempted to approximate the user's current state to adapt a game's difficulty or simulation. The user's state evaluation methods come in two main categories: subjective and objective~\cite{halbig2021systematic}. In the subjective method, the subjects are explicitly asked to express their experience through interviews, questionnaires, and think-aloud paradigms~\cite{greenwald2017investigating}. Although several studies have employed this method, people often judge their internal state poorly~\cite{halbig2021systematic}. In contrast, objective methods do not need the user to evaluate their experience. Instead, they examine the user's behaviour by direct observation or analysis of physiological data; hence, they are potentially more promising than subjective methods.

Physiological measurements can be collected during VR immersion fairly unobtrusively and capture features that correlate with sub-conscious states~\cite{halbig2021systematic}. They provide high-quality quantitative data that can be utilized in Machine Learning (ML) approaches to predict arousal and valence, anxiety, stress, and cognitive workload~\cite{halbig2021systematic}. Studies show that measuring stress/anxiety is valuable in designing adaptive and personalized VRET systems~\cite{halbig2021systematic}.


There are different use cases of physiological measurements in VR, which come in four main categories: therapy and rehabilitation, training and education, entertainment, and general or functional VR properties \cite{halbig2021systematic}. Physiological measurements in therapy and rehabilitation applications help assess the effectiveness of therapy \cite{kothgassner2019virtual, shiban2017diaphragmatic}, adapting the therapeutic system \cite{bualan2020investigation}, and offering appropriate feedback \cite{blum2019heart}. There are two main methods for adapting therapeutic systems: rule-based and machine learning. 

The rule-based studies use a threshold to predict a user's stress, or anxiety levels \cite{kritikos2021personalized}, e.g., considering the user's heart rate variability below/more than 100 bpm as a normal/abnormal states \cite{herumurti2019overcoming}. However, although these rule-based methods are fast and straightforward, they might not be a good representative of the actual user's state because the process requires expert knowledge, which is burdensome and subjective. Therefore, there is a need for more sophisticated techniques like machine learning approaches that automatically find a pattern in physiological signals to predict the user's stress or anxiety level.

VRET has increasingly been used to treat various anxiety disorders, including specific phobias, e.g., arachnophobia.
Adaptive VRET adapts the system's functionalities based on the subject's stress or anxiety. 
A handful of studies use machine learning methods for stress/anxiety classification in VR, as shown in Tabel~\ref{tab:table1}.

\begin{table*}[tb]
\centering
\caption{Studies that detect stress or anxiety in VR using ML methods.}
\label{tab:table1}
\begin{tabular}{|p{0.6cm}|p{4.9cm}|p{0.8cm}|p{0.9cm}|p{2.2cm}|p{0.95cm}|p{1.6cm}|p{1.35cm}|p{1.05cm}|}
\hline
 Study & Stimuli & Classes & Subjects & Biosignals & Window size & ML Method & Cross-Validation & Accuracy \\ \hline
 \cite{cho2017detection} & Arithmetic subtraction & 5 & 12 & PPG, EDA, SKT & 6 sec & K-ELM & LOOCV & 95\% \\ \hline
  \cite{ishaque2020physiological} & Virtual roller coaster + Stroop task & 2 & 14 & ECG, EDA, RESP & N/A & GB & 5-Fold & 85\% \\ \hline
  \cite{ham2017discrimination} & A guard patrolling in a dark room & 3 & 6 & PPG & N/A & LDA & 10-Fold & 79\% \\ \hline 
  \cite{tartarisco2015neuro} & Stressful work scenario for nurses & 4 & 18 & ECG, RESP & 5 min & SOM+FRBSs & 10-Fold/ LOSO & 83\% \\ \hline 
  \cite{robitaille2019increased} & Hitting moving targets & 2 & 12 & HR, Motion & N/A & DT & 2-Fold & 81.3\% \\ \hline 
  \cite{ahmad2021multi} & Virtual roller coaster & 3 & 9 & ECG & 1 sec & CNN+SVM & LOSO & 66.6\% \\ \hline
  \cite{hu2018research} & Standing on the ground and a plank & 4 & 60 & EEG, EOG & 20 sec & CNN & 10-Fold & 88.77\% \\ \hline
  \cite{wang2018combining} & Standing on the ground and a plank & 3 & 76 & EEG, EOG & N/A & SVM & 5-Fold & 96.2\% \\ \hline 
  \cite{kaur2019using} & Body leaning task on elevated ground & 2 & 10 & EEG, EOG & 10 sec & KNN & 5-Fold & 85\% (F1) \\ \hline
  \cite{balan2019automatic} / \cite{bualan2020investigation} & Exposure to different heights & 4 / 2 & 4 / 8 & EEG, PPG, EDA & N/A & DNN & Test-Train & 41.9\% / 89.5\% \\ \hline
  \cite{vsalkevicius2019anxiety} & Public speaking & 4 & 30 & PPG, EDA, SKT & 20 sec & SVM & 10-Fold/ LOSO & 86.3\% / 80.1\% \\ \hline
\end{tabular}

CNN:  Convolutional Neural Network, DNN: Deep Neural Network, DT: Decision Tree, FRBSs: Fuzzy rule-based module, GB: Gradient Boost, K-ELM: Kernel-based Extreme-learning machine, KNN: K Nearest Neighbor, LDA: Linear Discriminant Analysis, SOM: Self-Organizing map,  SVM: Support Vector Machine. 
ECG: Electrocardiogram, EDA: Electrodermal activity, EEG: Electroencephalography, EOG: Electrooculography, HR: Heart Rate, PPG: Photoplethysmogram, RESP: Respiration, SKT: Skin Temperature.

\end{table*}

Several stress/anxiety detection studies~\cite{cho2017detection,ishaque2020physiological,ham2017discrimination,tartarisco2015neuro,robitaille2019increased,ahmad2021multi} have used a dynamic or unpredictable virtual environment to induce stress in users, e.g., a roller-coaster ride~\cite{ishaque2020physiological}, or a guard patrolling in a dark and gloomy room~\cite{ham2017discrimination}. These environments are often combined with an additional assignment, e.g., an arithmetic task \cite{cho2017detection} or a Stroop task~\cite{ishaque2020physiological} to further trigger stress in subjects. Although these studies are valuable in predicting stress levels, they do not apply to adaptive VRET. The type of stress that VRET should be focused on must adversely affect the individuals' quality of life, such as avoiding outdoor activities because of arachnophobia, where spiders may be present.



Other studies have focused on fear of heights~\cite{hu2018research, kaur2019using, wang2018combining, bualan2020investigation, balan2019automatic}, or speech in front of a crowd \cite{vsalkevicius2019anxiety}. Although the studies have focused on valuable types of stress/anxiety for VRET, they mostly used Electroencephalography (EEG), which requires head-mounted sensors that are expensive and intrusive. Instead of using EEG, Salkevicius et al.~\cite{vsalkevicius2019anxiety} suggested using PPG, EDA, and SKT, which are less invasive. Nevertheless, according to ~\cite{long1982effect}, heart-rate, and blood pressure significantly increase while speaking, suggesting that the changes might not correspond to the amount of stress/anxiety alone. Instead, relaxing/stressful states should be considered in the data collection protocol to make sure changes in physiological measures are mainly associated with anxiety/stress. In addition, the validation method of several studies~\cite{cho2017detection, ishaque2020physiological, ham2017discrimination, robitaille2019increased, hu2018research, wang2018combining, kaur2019using, balan2019automatic, bualan2020investigation} cannot evaluate the system's performance in real-world scenarios. To obtain an independent subject score corresponding to more realistic results for real-life deployment, leave-one-subject-out (LOSO) cross-validation should be applied, i.e., \cite{ahmad2021multi, tartarisco2015neuro, vsalkevicius2019anxiety}. Therefore, estimating subjects' stress/anxiety using physiological signals in scenarios suitable for VRET has remained challenging. 

In this study, we design a virtual reality protocol appropriate for VRET to collect a physiological dataset via a Photoplethysmography (PPG) biosignal sensor, which is widely available and non-intrusive.
We also examine the capability of a stressful virtual environment to induce stress in subjects. The protocol consists of two VR environments: relaxing and stressful. First, we collected the PPG data from sixteen healthy subjects exposed to our VR environments. Then we apply classification methods to learn a model to automatically detect the user's state (relaxed or stressed). The contributions of our work are: 1) developing a novel, stressful VR environment meant to support arachnophobia therapy, and 2) demonstrating the ability to accurately detect stress/anxiety via PPG, a widely available sensor, which allows for more general applications

\section{Methodology}

\begin{figure*}[t]
  \centering
  \begin{subfigure}[b]{0.3\textwidth}
    \includegraphics[width=\textwidth]{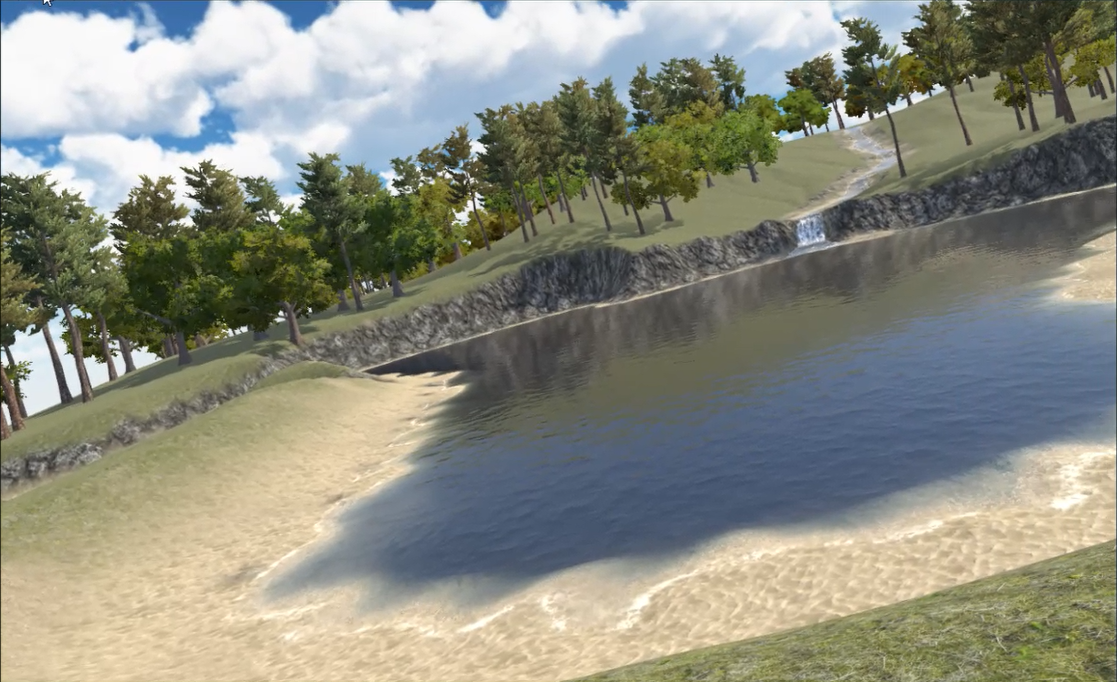}
     \caption{Relaxing} 
     \label{fig: relaxing}
  \end{subfigure}
  \begin{subfigure}[b]{0.3\textwidth}
    \includegraphics[width=\textwidth]{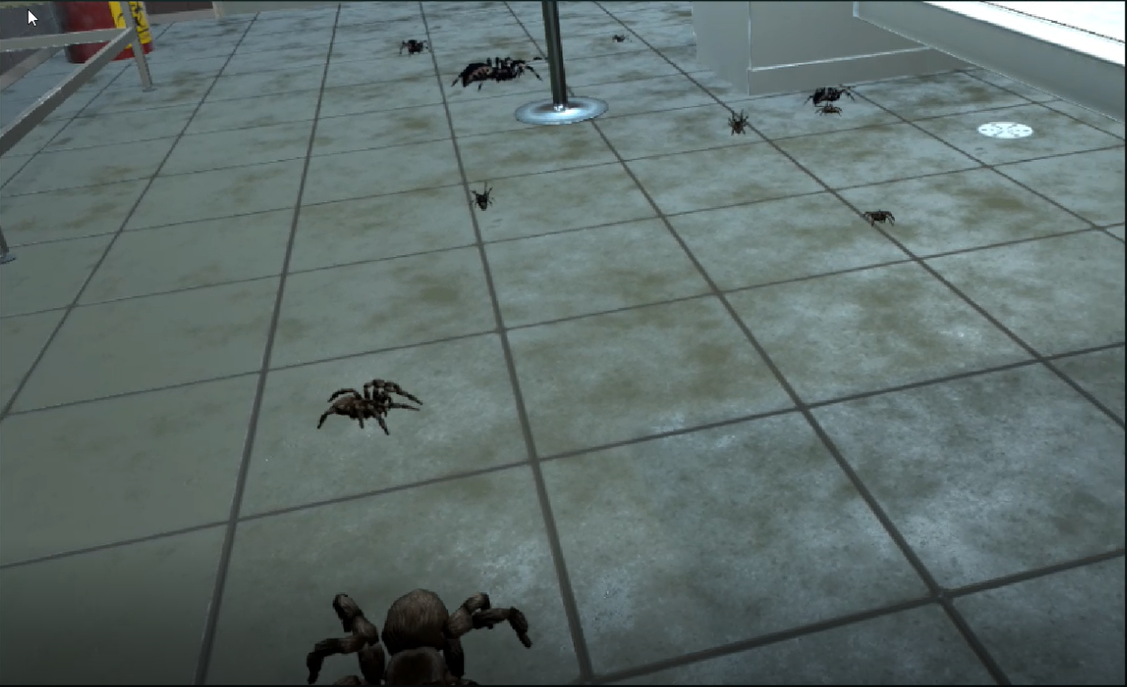}
    \caption{Stressful}
    \label{fig: stressful}
  \end{subfigure}
  \begin{subfigure}[b]{0.3\textwidth}
    \includegraphics[width=\textwidth]{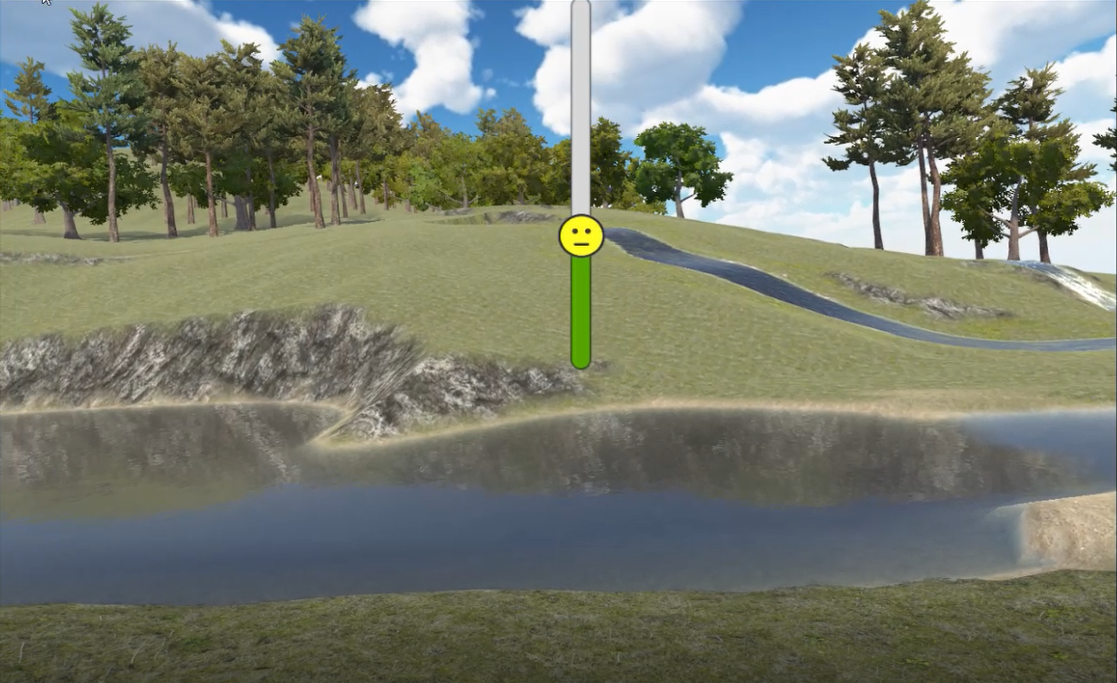}
    \caption{SUDs rating}
    \label{fig: rating}
  \end{subfigure}
  \caption{Developed VR environment.}
  \label{fig: VR}
\end{figure*}

\subsection{Virtual Environments}
We develop two VR environments: a relaxing and a stressful environment, shown in Fig~\ref{fig: relaxing} and Fig~\ref{fig: stressful}, respectively. The relaxing environment presents nature, including mountains, rivers, waterfalls, and trees with bird sounds. The stressful environment shows a morgue with scary sounds, including spiders of different sizes and colours. Spiders either move randomly or follow the participant. In either case, they always maintain a one-meter distance from the participant. We develop the VR environments using the Oculus Rift S~\footnote{https://www.oculus.com/rift/} head-mounted display system and Unity3D.    

\subsection{Participants}
Sixteen healthy subjects (9 Males, 7 Females) aged 18-35 participated in our experiment. We used a Fear of Spiders Questionnaire~\cite{szymanski1995fear} to ensure that the subjects did not have arachnophobia.   
All participants provided their informed consent to undertake the experiment by reading and signing a form explaining the background, objectives and procedures of the study and the confidential handling of all the collected data. Participants received a \$10 gift card for their participation.

\subsection{Measures}


According to the literature, heart rate variability (HRV) changes in response to stress~\cite{kim2018stress}. HRV can be derived from either ECG or PPG. Although ECG has been effectively used for measuring HRV, it is also adequate to use finger PPG during rest and mild mental stress~\cite{singstad2021estimation}. Moreover, PPG can be obtained via low-cost, non-invasive and portable methods, i.e., smartwatches \cite{lee2019can} that make this approach more accessible and easy to use.
We collected PPG signals from the participants during the experiment using a Bitalino~\footnote{https://www.pluxbiosignals.com/products/psychobit} device by attaching a single sensor to their left index finger. 
We obtain the ground truth, i.e., the amount of stress/anxiety during the VR experiences, using the widely used Subjective Unit of Distress scale (SUDs)~\cite{mccabe2015subjective}. The scale measures from 0 to 100, where 0 means total relaxation and 100 represents the highest anxiety ever felt. 

The participants were asked to fill out questionnaires after the experiment. They included demographic questions (age, gender, experience in playing video games and VR), a Simulator sickness questionnaire (SSQ)~\cite{bimberg2020usage} to quantify sickness elicited by the VR system, and the State-Trait Anxiety Inventory (STAI)~\cite{tluczek2009support} to measure self-assessment anxiety in each environment.  

\subsection{Procedure}
The whole procedure of our experiment was as follows. First, a participant signed an informed consent form. Second, we attached the PPG sensor to their left index finger and started recording their data. Then, they were asked to wear a head-mounted display and stand throughout the experiment (they were not allowed to move to eliminate the movement's effect on physiological recordings). They were then exposed to a trial environment (2 minutes) to get familiar with the controllers and practice the point and teleport locomotion~\cite{bozgeyikli2016point}. Afterwards, they were exposed to relaxing and stressful environments for seven minutes each (14 minutes total). They could either stay in their position or explore the environments using the point and teleport method.
The total participants' interaction with the VR environments was less than 20 minutes to minimize the chances of the participants experiencing general symptoms, e.g., general discomfort, fatigue, and dizziness, as suggested by~\cite{howarth1997occurrence}.  
Finally, each participant evaluated their SUDs through a VR interface every 2 minutes (Fig~\ref{fig: rating}). They were asked to fill out questionnaires after the experiment.      

\section{Evaluation and Results}

\begin{table*}[tb]
\centering
\small
\caption{The accuracy of different machine learning algorithms for test data on our collected dataset.}
\label{Table: ML}
\begin{tabular}{|p{1.2cm}|p{1.1cm}|p{0.9cm}|p{1.1cm}|p{0.9cm}|p{1.1cm}|p{1.1cm}|p{1.1cm}|p{0.9cm}|p{1.1cm}|}
\hline
 & \rotatebox[origin=c]{90}{{\parbox[c]{2cm}{\centering Support Vector Machine}}} & \rotatebox[origin=c]{90}{{\parbox[c]{2cm}{\centering Random Forest}}} & \rotatebox[origin=c]{90}{{\parbox[c]{2cm}{\centering AdaBoost (Decision Tree)}}} & \rotatebox[origin=c]{90}{{\parbox[c]{1.5cm}{\centering XGBoost}}} & \rotatebox[origin=c]{90}{{\parbox[c]{2cm}{\centering Neural Network (MLP 2 layers)}}} & \rotatebox[origin=c]{90}{{\parbox[c]{2cm}{\centering Neural Network (MLP 4 layers)}}} & \rotatebox[origin=c]{90}{{\parbox[c]{2cm}{\centering Linear Discriminant Analysis}}} & \rotatebox[origin=c]{90}{{\parbox[c]{2cm}{\centering K-Nearest Neighbour}}} & \rotatebox[origin=c]{90}{{\parbox[c]{2cm}{\centering Stochastic Gradient Descent}}} \\ \hline
Accuracy & 64.04\% & 58.56\% & 61.82\% & 59.56\% & 63.13\% & 59.19\% & \textbf{70.56\%} & 53.40\% & 64.67\% \\ \hline
\end{tabular}
\end{table*}

\begin{figure}[t]
  \centering
  \includegraphics[width=0.8\linewidth]{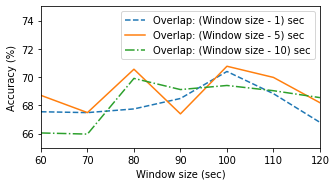}
  \caption{The accuracy of the best classification methods for each fixed-size sliding windows with different overlaps}
  \Description{The orange line shows the accuracy for different sliding windows for 5 seconds shifting, e.g., window size 100 with 95 seconds overlaps. }
  \label{fig: Windows}
  \vspace{-5mm}
\end{figure}

We formulate the problem of stress/anxiety detection as a binary classification problem where class labels are the subjects' \quotes{relaxing} and \quotes{stressful} states.
First, we segment the PPG signal by a fixed-size sliding window, i.e., 60, 70, 80, 90, 100, 110 and 120 seconds (the results are shown in Fig \ref{fig: Windows}). We choose an 80-sec window size with 75-sec overlaps since it yields the most accurate result over other window sizes.    
We apply a third-order Butterworth bandpass filter to remove artifacts and noises at 0.5Hz-8Hz. 
Then we use the Neurokit python package~\cite{Makowski2021neurokit} to extract HRV features that are suggested in~\cite{pham2021heart}: time-domain (e.g., RMSSD, MeanNN, SDNN, SDSD), frequency domain (e.g., Spectral power density in various frequency bands, Ratio of LF to HF power, Normalized LF and HF), and non-linear domain (e.g., Spread of RR intervals, Cardiac Sympathetic Index). Based on ~\cite{pham2021heart}, extracting frequency domain indices requires at least 60 seconds of recording; that is why window sizes less than 60 seconds are not considered.

We consider the 35 highest score features based on ANOVA F-value for classification. The five best features include:
Median absolute deviation of the RR intervals divided by the median of the absolute differences of their successive differences (MCVNN), Shannon entropy of HRV (ShanEn), Proportion of successive NN interval differences larger than 20 ms (pNN20), Standard deviation of the RR intervals divided by mean RR intervals (CVNN), and Interquartile range of the RR intervals (IQRNN).


We apply leave-one-subject-out (LOSO) as our cross-validation method, i.e., consider the data of one subject as a test and the remaining data as training, repeating it for every subject, and report the average accuracy (equivalent to a 16-fold cross validation). The results of running different machine learning algorithms are shown in Table~\ref{Table: ML}. Our results show that Linear Discriminant Analysis (LDA) can classify at 70.56\% accuracy.
This result is about as expected, considering related works that apply LOSO and only use HRV features via PPG or ECG to detect stress~\cite{ahmad2021multi, behinaein2021transformer}.

We also analyze the self-reported SUDs through each VR environment using the two-tailed Mann–Whitney U test. The results show that the SUDs are significantly higher in the stressful environment ($p_{value}<0.00001$), proving that the VR environment could induce stress in the subjects.

Based on our questionnaires, all sixteen subjects found the first environment more relaxing than the second environment. Fifteen subjects found the second one more stressful, except one subject who found both environments stressful. 

Subjects felt involved in the relaxing and stressful environment with $5.62\pm1.31$ and $5.5\pm1.27$ out of 10 scores (10 being the complete immersion), respectively. This result indicates that the subjects felt roughly the same immersion in both environments.
The subjects considered the potential of using this technology in therapeutic content as $5.75\pm1.08$ out of 10.

The subjects indicated, in an open-ended question, that the insects/bugs' features that scare them most were: movement (8 out of 16 subjects), size (7 out of 16 subjects), speed (3 out of 16 subjects), hairiness (3 out of 16 subjects), jumping (2 out of 16 subjects), wings (2 out of 16 subjects), lots of eyes (1 out of 16 subjects), and distance to them (1 out of 16 subjects). 
This shows that the movement and size are the most important features that induce stress in subjects.

Based on the questionnaire, the unwanted adverse side effects of the immersion in VR subjects include general discomfort (6 out of 16 subjects), eye strain (2 out of 16 subjects), blurred vision (2 out of 16 subjects), sweating (2 out of 16 subjects), dizziness with eyes open (1 out of 16 subjects), nausea (1 out of 16 subjects), difficulty concentrating (1 out of 16 subjects), and the fullness of head (1 out of 16 subjects).
This reveals that, although only one subject felt multiple side effects, most of the subjects were comfortable while using VR. Nevertheless, a handful of subjects felt general discomfort, which is typical while using VR.

\section{Discussion}


We developed a platform for inducing stress/anxiety, suitable for VRET studies, specifically arachnophobia. Our platform produces meaningful physiological data that can be used to detect users' stress/anxiety states accurately. 

In comparison to related work, our platform can be widely applied to adaptive VRET studies. It performs accurately using HRV features via PPG signals only. Moreover, PPG is widely available, e.g., smartwatches, non-intrusive and cost-effective. We also demonstrate generality via evaluating the platform on new users (LOSO).

It is important to note that we tested our platform on non-arachnophobic individuals who are less sensitive to spiders. However, the final users of the system will be arachnophobic people who are expected to show more significant physiological responses to spider stimuli.

One limitation of our platform is that we formulate the problem as a binary classification problem where class labels are the subjects' \quotes{relaxing} and \quotes{stress} states. It could be problematic if inducing a particular dosage of stress/anxiety in subjects is desired in VRET. In order to overcome this limitation, it can be assumed that the membership probability of a \quotes{stress} state of each sample represents their stress/anxiety level. Then, rounding the estimated stress/anxiety to one decimal place smoothes out the prediction noise. For example, if a sample's membership probabilities are 0.18 and 0.82 for \quotes{relaxing} and \quotes{stress} states, respectively, we can assume a stress/anxiety level equal to 0.8. 
However, corresponding ground truth data does not exist for validating this approach. 

Another drawback of our platform is the interaction of the users with the relaxing environment. We observe that some users tend to explore the relaxing environment such as jumping/submerging into water, climbing hills to find cliffs and finding the \quotes{edges of the world}. These interactions can also induce stress/anxiety in users. These interactions can be prevented by redesigning the environment and/or limiting interaction methods (e.g., lying on a poolside bench). We would then expect to obtain more accurate results. Nevertheless, these changes should be applied carefully to prevent boredom in users.

Our proposed platform can be used as a component in adaptive VRET systems, wherein a content generator, e.g., a Reinforcement Learning (RL) Agent~\cite{mahmoudi2021automated}, gets as input stress/anxiety level prediction and changes the content of the virtual environment suitable for the exposure therapy, e.g., spiders' attributes. These adaptive VRET systems are shown to be more effective~\cite{zahabi2020adaptive} than non-adaptive systems because the content generator personalizes the VR environment based on each individual's need. We plan to use our platform in adaptive VRET systems and examine if the content generator can effectively change the VR environment based on users' stress/anxiety levels.

\section{Conclusion}

We present a model to detect anxiety/stress automatically in VR using widely available PPG signals. This model is valuable for designing adaptive VRET systems that refine their functionalities based on the user's state. Our best classification model could predict relaxing or stressful states of subjects (binary classification) with a 70.56\% accuracy using LOSO cross-validation, opening the door for live VRET applications.
\bibliographystyle{ACM-Reference-Format}
\bibliography{my_bib}

\end{document}